\documentclass{article}

\usepackage{arxiv}
\usepackage[utf8]{inputenc} 
\usepackage[T1]{fontenc}    
\usepackage{hyperref}       
\usepackage{url}            
\usepackage{booktabs}       
\usepackage{amsfonts}       
\usepackage{nicefrac}    

\usepackage{microtype}      
\usepackage{graphicx}
\usepackage{amsmath}
\usepackage{enumerate}
\usepackage{enumitem}
\usepackage{natbib}
\usepackage{doi}
\usepackage{todonotes}
\usepackage{xcolor}
\usepackage{tcolorbox}
\tcbuselibrary{breakable,skins}
\usepackage{graphicx}
\usepackage{subcaption}  
\usepackage{adjustbox}   
\usepackage{listings}  
\usepackage{xcolor}   
\usepackage{geometry}
\usepackage[utf8]{inputenc}
\usepackage{tcolorbox}
\usepackage{mdframed}
\usepackage{framed}
\usepackage{xcolor}
\usepackage{listings}
\usepackage{paracol}
\usepackage{fancybox}

\definecolor{lightblue}{RGB}{230,240,255}
\definecolor{lightgreen}{RGB}{230,255,230}
\definecolor{lightyellow}{RGB}{255,250,230}
\definecolor{lightpink}{RGB}{255,230,240}

\definecolor{brightblue}{RGB}{0, 122, 255}
\usepackage{fontawesome}

\newcommand{\githubinfo}{%
  \vspace{-0.7em}
  \begin{center}
    \href{https://github.com/AQ-MedAI/MrlX}%
         {\faGithub~Code and Dataset: \texttt{https://github.com/AQ-MedAI/MrlX}}
  \end{center}
  \vspace{-0.7em}
}

\title{%
  \raisebox{-0.35ex}{\includegraphics[width=0.8cm]{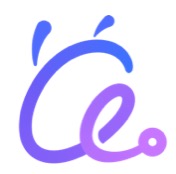}}%
  \hspace{3.5px}\emph{Multi-Agent Deep Research}: Training Multi-Agent Systems with M-GRPO
}

\author{
Haoyang Hong$^{2}$\thanks{Work done during an internship at Ant Group.} \quad
Jiajun Yin$^{1}$ \quad
Yuan Wang$^{1}$ \quad
Jingnan Liu$^{1}$ \quad
Zhe Chen$^{1}$ \quad
Ailing Yu$^{1}$ \\
Ji Li$^{1}$ \quad
Zhiling Ye$^{1}$ \quad
Hansong Xiao$^{1}$ \quad
Yefei Chen$^{1}$ \quad
Hualei Zhou$^{1}$ \quad
Yun Yue$^{1}$ \\
Minghui Yang$^{1}$ \quad
Chunxiao Guo$^{1}$ \quad
Junwei Liu$^{1}$ \quad
Peng Wei$^{1}$ \quad
Jinjie Gu$^{1}$ \\[1ex]
$^{1}$Ant Group \qquad
$^{2}$Imperial College London
}

\renewcommand{\shorttitle}{A preprint}

\begin{document}
\maketitle

\githubinfo
\begin{center}
\begin{abstractbox}
Multi-agent systems perform well on general reasoning tasks. However, the lack of training in specialized areas hinders their accuracy. Current training methods train a unified large language model (LLM) for all agents in the system. This may limit the performances due to different distributions underlying for different agents. Therefore, training multi-agent systems with distinct LLMs should be the next step to solve. However, this approach introduces optimization challenges. For example, agents operate at different frequencies, rollouts involve varying sub-agent invocations, and agents are often deployed across separate servers, disrupting end-to-end gradient flow. To address these issues, we propose M-GRPO, a hierarchical extension of Group Relative Policy Optimization designed for vertical Multi-agent systems with a main agent (planner) and multiple sub-agents (multi-turn tool executors). M-GRPO computes group-relative advantages for both main and sub-agents, maintaining hierarchical credit assignment. It also introduces a trajectory-alignment scheme that generates fixed-size batches despite variable sub-agent invocations. We deploy a decoupled training pipeline in which agents run on separate servers and exchange minimal statistics via a shared store. This enables scalable training without cross-server backpropagation. In experiments on real-world benchmarks (e.g., GAIA, XBench-DeepSearch, and WebWalkerQA), M-GRPO consistently outperforms both single-agent GRPO and multi-agent GRPO with frozen sub-agents, demonstrating improved stability and sample efficiency. These results show that aligning heterogeneous trajectories and decoupling optimization across specialized agents enhances tool-augmented reasoning tasks.
\end{abstractbox}
\end{center}

\section{Introduction}
\vspace{-0.45em}

AI agent systems have demonstrated significant potential in solving complex tasks by integrating large language models (LLMs). Early single-agent approaches, which combine reasoning and actions, have improved robustness and ability in open-domain tasks \citep{yao2022react, schick2023toolformer, qin2023toolllmfacilitatinglargelanguage, patil2024gorilla, xu2023rewoodecouplingreasoningobservations}. However, as evaluations move from static NLP benchmarks to real-world scenarios, especially in specialized domains like medicine, performance gains are inconsistent. While single agent systems perform well on short tasks, they struggle with long-horizon coordination, compounding errors, and dynamic environments \citep{zhou2024webarenarealisticwebenvironment, liu2025agentbenchevaluatingllmsagents}.

\begin{figure}[ht]
    \centering
    \includegraphics[width=0.95\linewidth]{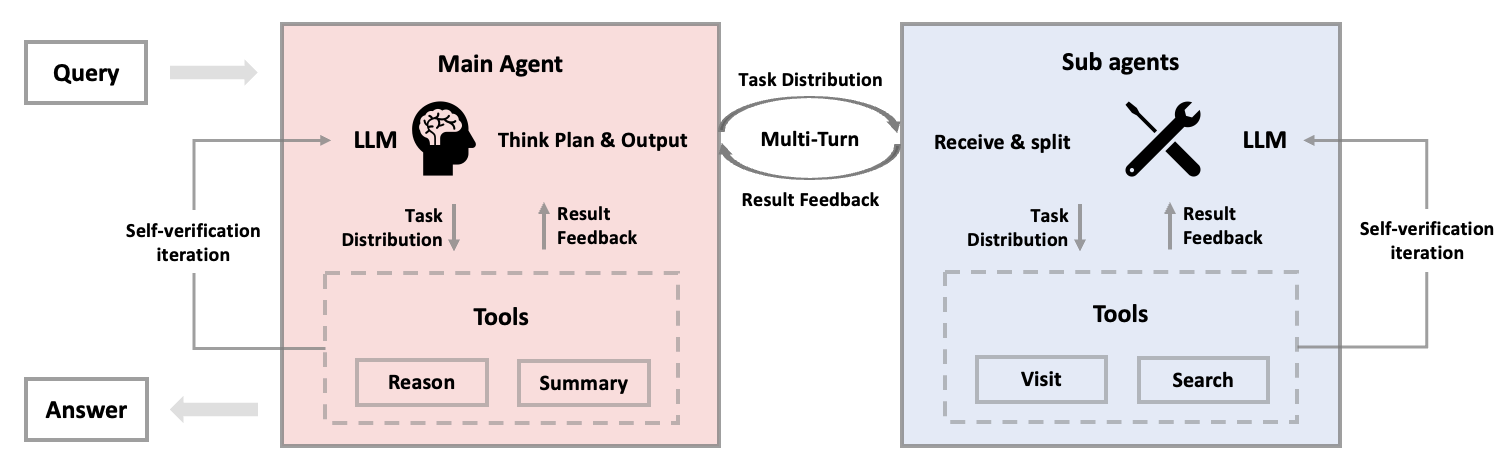}
    \vspace{0.25em}
    \caption{\textbf{System workflow with coordinated main and sub-agents.}
  A user query is fed to the main agent \(\mathcal{M}\), which plans, reasons, and delegates subtasks to specialized sub-agents \(\{\mathcal{S}_i\}\) if needed (e.g., visit/browsing and search tools).
  Sub-agents return structured feedback to \(\mathcal{M}\), which integrates evidence, performs verification, and produces the final answer. 
  Both \(\mathcal{M}\) and \(\mathcal{S}_i\) may iterate via self-verification loops before feedback outputs to $\mathcal{M}$.}
    \label{fig:workflow}
    \vspace{-0.25em}
\end{figure}

Real-world tasks often require decomposition, critique, and iterative processes, which a single agent cannot handle effectively. Benchmarks based on functional websites and multi-environment suites highlight common failures: brittle long-term plans, poor self-checking, and challenges in task delegation \citep{zhou2024webarenarealisticwebenvironment, liu2025agentbenchevaluatingllmsagents}. These limitations drive the need for multi-agent architectures, where agents assume distinct roles (e.g., planner, researcher, solver) and verify each other's actions \citep{wu2023autogenenablingnextgenllm, li2023camelcommunicativeagentsmind, du2023improvingfactualityreasoninglanguage}. Vertical architectures, where one agent leads and others report to it, outperform horizontal systems with equal agent roles. Studies show that multi-agent systems with a leader complete tasks nearly 10\% faster than those without \citep{masterman2024landscapeemergingaiagent}. Therefore, we adopt a vertical architecture for improved performance. The work flow is illustrated in Figure~\ref{fig:workflow}.

While multiple agents improve coverage and reliability, it still requires sophisticated training for complex tasks or specific domains. However, training multi-agent systems with different LLMs intergrated is challenging. The main issue in vertical multi-agent systems is the imbalance in rollout numbers between the leader and task agents, leading to unstable and asynchronous training. Additionally, deploying different LLMs for each agent across separate training servers complicates gradient flow, making standard backpropagation methods unworkable.

To address these challenges, we propose \textbf{M-GRPO}, a reinforcement-learning framework for training separate LLMs as distinct agents in vertical multi-agent systems. Our contributions include:
\vspace{-0.5em}
\begin{itemize}[leftmargin=12pt]
\setlength\itemsep{-0.25em}
    \item \textbf{Vertical multi-agent RL formalization}.
    We formalize training for a vertical multi-agent architecture where a main agent (\(\mathcal{M}\)) plans and sub-agents (\(\mathcal{S}\)) execute tool-use subtasks. We extend Group Relative Policy Optimization to this nested setting by computing group-relative advantages that respect hierarchical credit assignment.
    \item \textbf{Trajectory-alignment for variable sub-invocations}.
    We introduce a simple but effective alignment scheme to handle the variable number of \(\mathcal{S}\) invocations per rollout. By choosing a target \(D_{\max}\) and masking (or duplicating/dropping) sub-trajectories, we obtain fixed-shape batches without destabilizing the group baseline, enabling efficient, batched policy-gradient updates.
    \item \textbf{Empirical gains on real-world agent benchmarks.}
    On GAIA, XBench-DeepSearch and WebWalkerQA benchmarks, our method yields consistent improvements over a strong single-agent GRPO and a multi-agent system with frozen sub-agent baselines across training checkpoints.
\end{itemize}

\section{Related work}
\vspace{-0.45em}
\subsection{Reinforcement Learning for LLM Post-Training}
\vspace{-0.25em}
Reinforcement learning from human feedback (RLHF)~\citep{ouyang2022traininglanguagemodelsfollow} established the contemporary post-training pipeline for aligning large language models (LLMs). It often instantiates proximal policy optimization (PPO)~\citep{schulman2017proximalpolicyoptimizationalgorithms} with a Kullback–Leibler (KL) divergence penalty. However, value functions and costly rollouts motivated simpler preference fitting. Accordingly, Direct Preference Optimization (DPO)~\citep{rafailov2023dpo} and RRHF~\citep{yuan2023rrhfrankresponsesalign} remove explicit RL while keeping preference alignment. More recently, group-relative objectives such as Group Relative Policy Optimization (GRPO)~\citep{shao2024deepseekmath} have shown strong gains in mathematical and program reasoning. DeepSeek-R1 demonstrated that pure-RL can incentivize longer, higher-quality reasoning traces without explicit human-labeled trajectories~\citep{guo2025deepseekr1}. In parallel, work on language feedback and reflection mechanisms complements preference learning by providing process-level signals rather than outcome-only labels~\citep{shinn2023reflexionlanguageagentsverbal}. Taken together, these advances deepen planning and improve reliability of LLMs. As capability rises, AI agent, as an important application mode of LLM, starts to show its extraordinary performances in dealing with multi-step tasks. The evolution of agentic systems has progressed from single-agent architectures to more sophisticated multi-agent frameworks, enabling the collaborative resolution of increasingly complex and high-level goals. 

\subsection{Single-Agent system training}
\vspace{-0.25em}
Early single-agent systems emphasized planning and thinking without changing model parameters. ReAct~\citep{yao2022react} and Self-Ask~\citep{press2023measuringnarrowingcompositionalitygap} structure reasoning–acting–decomposition for complex tasks. Tree of thought (ToT) and Graph of thought (GoT) explore and select among branched thoughts~\citep{yao2023treethoughtsdeliberateproblem, Besta_2024}. 
To improve reliability, self-monitoring approaches introduce process-level checks and feedback, including Reflexion-style reflection and LLM-as-a-judge techniques~\citep{shinn2023reflexionlanguageagentsverbal, gu2025surveyllmasajudge}. 

To further improve the performance of single-agent system, researches also focus on how to fine-tune single-agent system. Single-agent training has progressed from supervised fine-tuning and RLHF to lighter direct preference objectives (DPO/RRHF), process-level rewards (PRMs~\citep{lightman2023letsverifystepstep}), and reasoning-targeted RL (GRPO). Aside from aligning LLMs to expected answer formats, other works focus on training single-agent to perform proper and better tool calling~\citep{schick2023toolformerlanguagemodelsteach, patil2023gorillalargelanguagemodel}. Despite these advances, a single model often struggles with long-horizon tasks and heterogeneous skills. These limitations motivate a shift toward multi-agent systems that decompose roles to better match the growing complexity of real-world problems.



\subsection{Multi-agent system training}
\vspace{-0.25em}
Multi-agent systems (MAS) have shown inspiring collective intelligence even in inference-time collaboration~\citep{guo2024largelanguagemodelbased}. Lot of works propose sophisticated frameworks for MAS and shows incredible performances~\citep{chen2023agentversefacilitatingmultiagentcollaboration, wu2023autogenenablingnextgenllm, hong2024metagptmetaprogrammingmultiagent, qian2024chatdevcommunicativeagentssoftware}. However, when it comes to complex tasks or tasks in specific domains such as medicine, MASs still require training for better performance. Recent work trains collaborative agents end-to-end. For instance, Debate/self-play have been used to produce supervisory signals without ground-truth labels, improving judge accuracy and reasoning quality~\citep{arnesen2024traininglanguagemodelswin,chen2024selfplayfinetuningconvertsweak,kenton2024scalableoversightweakllms, xue2025comascoevolvingmultiagentsystems}. Surveys synthesize these directions and open issues around coordination, evaluation, and reward shaping in LLM-based MASs~\citep{chen2025surveyllmbasedmultiagentsystem,luo2025largelanguagemodelagent,sun2024llmbasedmultiagentreinforcementlearning}. 

Recent works introduce RL training frameworks for multi-agent system. \cite{bo2024reflective} introduces agent-level rewards and counterfactual credit assignment tailored to LLM agents. \textsc{MALT}~\citep{motwani2025maltimprovingreasoningmultiagent} organizes generator--verifier--refiner roles and propagates credit across a pipeline to improve math and commonsense reasoning. \textsc{MAPoRL}~\citep{park2025maporlmultiagentpostcotrainingcollaborative} co-trains agents with multi-agent reinforcement learning to elicit cooperative behaviors via discussion rewards, reporting gains over single-agent post-training. The method adjusts PPO for MAS training. Hence, it can be costly with value function required. Further, \cite{mo2025multiagenttoolintegratedpolicyoptimization} and \cite{chen2025marsoptimizingdualsystemdeep} propose GRPO-based methods for MAS training. However, both of them share a single underlying LLM across agents, which ties parameters and inductive biases. This hinders role specialization when agents perform heterogeneous functions (e.g., plan vs. split task) with distinct state–action/data distributions, motivating role-specialized LLMs. \cite{mrlx2025} co-trains dual agents with distinct LLMs under a synchronous Q\&A loop where both take the same number of actions per rollout. In vertical systems, however, agents act at different frequencies, creating reward misalignment and unstable cross-agent credit assignment. We address this with \emph{Multi-agent Group Relative Policy Optimization} (M-GRPO), which aligns heterogeneous trajectories via group-relative advantages.

\section{Problem setup}
\vspace{-0.45em}
\subsection{Architecture setup}

\begin{figure}
    \centering
    \includegraphics[width=0.8\linewidth]{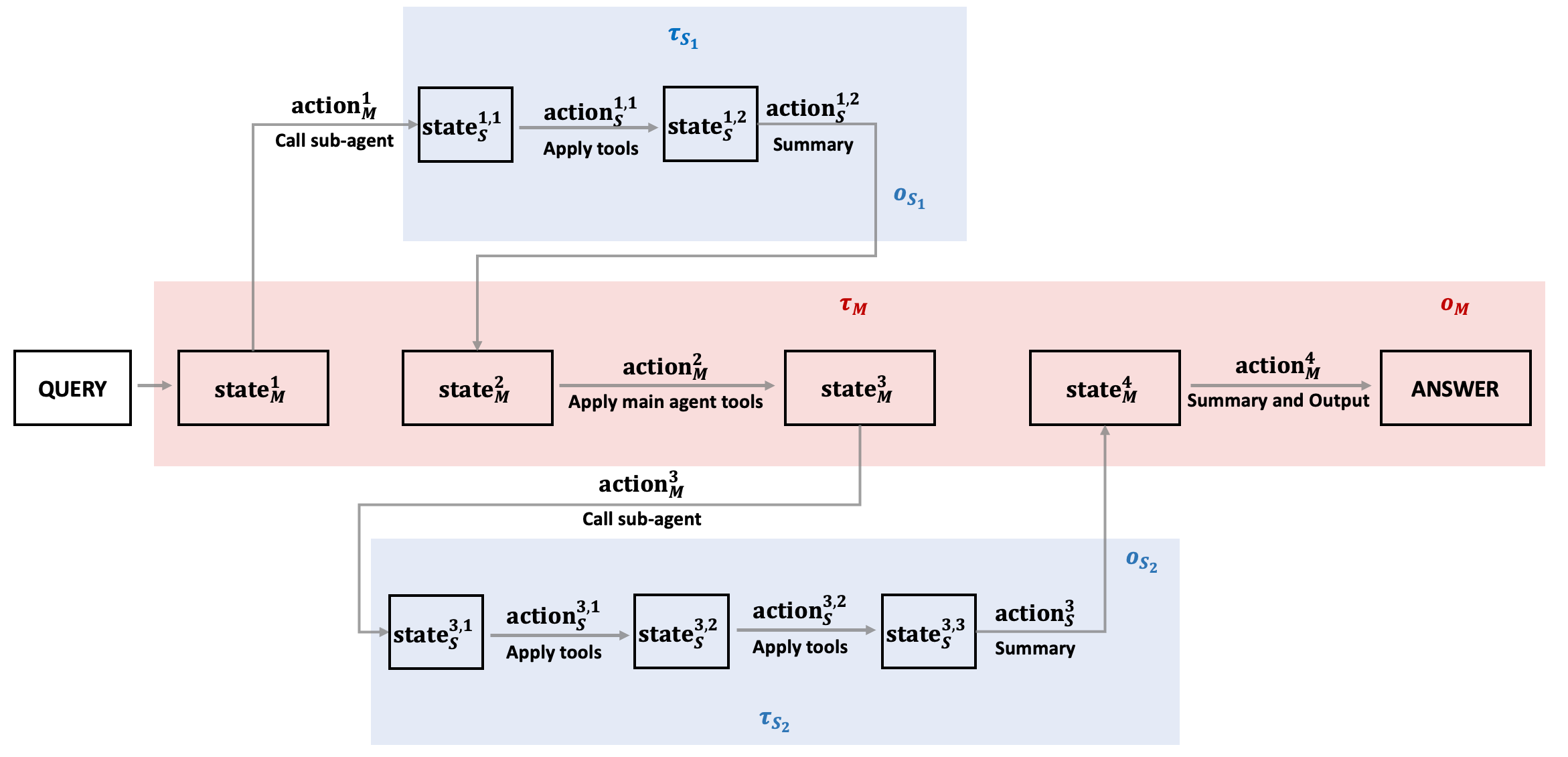}
    \caption{\textbf{One rollout with nested \(\mathcal{M}\!\to\!\mathcal{S}\) interactions.}
  The main agent \(\mathcal{M}\) follows trajectory \(\tau_{\mathcal{M}}\) (red) and may distribute subtasks by invoking the sub-agent \(\mathcal{S}\) multiple times (e.g., \(a_{\mathcal{M}}^{1}\) and \(a_{\mathcal{M}}^{3}\)).
  Each invocation generates a sub-trajectory \(\tau_{\mathcal{S}_i}\) (blue) that performs tool-use steps and returns a summarized message \(o_{\mathcal{S}_i}\) to \(\mathcal{M}\).
  The main trajectory integrates these intermediate results and finally outputs the answer \(o_{\mathcal{M}}\).}
    \label{fig:rollout}
\end{figure}

\vspace{-0.25em}
We employ a vertical two-agent architecture as the base case for hierarchical systems. A \textbf{main agent} \(\mathcal{M}\) plans, decomposes a user query \(q\) into subtasks, decides when to delegate, and synthesizes a final answer. It is also equipped with a reason tool for analyzing complex question. A \textbf{sub-agent} \(\mathcal{S}\) executes delegated subtasks that typically involve tool use (e.g., search, visit) and reports findings back to $\mathcal{M}$. Communication follows a call–return protocol: $\mathcal{M}$ issues a subtask specification and optional context to $\mathcal{S}$, $\mathcal{S}$ returns structured results (evidence, summaries, and metadata). This two-agent setting extends naturally to $1+n$ agents by instantiating a set \(\{\mathcal{S}_i\}_{i=1}^n\) with the same architecture but different functions. Without loss of generality, we focus on $n=1$ to present the core method cleanly in this paper (Figure.~\ref{fig:rollout} illustrates the workflow in one rollout).

\vspace{-0.25em}
\textbf{Action space}. 
Each agent’s policy comprises (i) \emph{token-level} textual emissions and (ii) \emph{discrete} control actions (tool invocation, delegation, stop). Text actions are trained at the token level; control actions are treated as single-step categorical choices interleaved in the trajectory.

\vspace{-0.25em}
\textbf{Notations}.
\label{sec:notation}
For one rollout in Figure.~\ref{fig:rollout}, the trajectory (including actions, states and rewards) and outputs are
\begin{align}
    &\tau = \big\{\tau_{\mathcal{M}}, \;\{\tau_{\mathcal{S}_i}\}_{i=1}^{d}\big\}\\
    &\tau_\mathcal{M}=\big\{a_{\mathcal{M},t}, \;s_{\mathcal{M},t}, \;r_{\mathcal{M},t}\big\}_{t=1}^{T_{\mathcal{M}}},\hspace{1em} \tau_{\mathcal{S}_i}=\big\{a_{\mathcal{S}_i,t},\; s_{\mathcal{S}_i,t}, \;r_{\mathcal{S}_i,t}\big\}_{t=1}^{T_{\mathcal{S}_i}}\\
    & o  = \big\{o_{\mathcal{M}}, \;\{o_{\mathcal{S}_i}\}_{i=1}^{d}\big\}
\end{align}
where $d$ is the number of sub-agent invocations made by $\mathcal{M}$ in this rollout. The value of $d$ varies across rollouts. At time step $t$ out of total step $T_\mathcal{M}$, the main agent trajectory ($\tau_\mathcal{M}$) consists of action ($a_{\mathcal{M},t}$), state ($s_{\mathcal{M},t}$) and reward ($r_{\mathcal{M},t}$). In practice, the reward $r_{\mathcal{M},t}$ is broadcasted from the reward of final output of the main agent. Notations are similar for the $i$-th invocation of subagent. The final answer to \(\mathcal{M}\) is \(o_{\mathcal{M}}\); each sub-invocation returns \(o_{\mathcal{S},i}\).
Agent policies are denoted by \(\pi_{\theta_{\mathcal{M}}}\) and \(\pi_{\theta_{\mathcal{S}}}\), mapping states to actions.

\vspace{-0.25em}
\subsection{Reward setup}
\vspace{-0.25em}
The main agent is evaluated based on the quality of its final output $o_\mathcal{M}$:
\begin{equation}
\label{eq:Rm}
\mathcal{R}_{\mathcal{M}}(o_{\mathcal{M}}) = 
\begin{cases}
\alpha_1 \cdot r_{\text{format}} + \alpha_2 \cdot r_{\text{correct}} & \text{valid format} \\
0 & \text{invalid format}
\end{cases}
\end{equation}
where:
\begin{itemize}[leftmargin=25pt]
\vspace{-0.75em}
\setlength\itemsep{-0.25em}
\item $r_{\text{format}}$: Binary reward for valid output structure (proper JSON, required fields).
\item $r_{\text{correct}}$: Accuracy reward comparing final answer to ground truth.
\item $\alpha_1$ and $\alpha_2$ are hyperparameters that balance these two features.
\end{itemize}

\vspace{-0.25em}
If the output is in invalid format, the trajectory gets zero reward, regardless of correctness. This strict policy ensures:
\begin{itemize}[leftmargin=25pt]
\vspace{-0.75em}
\setlength\itemsep{-0.25em}
\item Invalid formats cannot break downstream processing in real deployments.
\item Partial correctness with invalid format is penalized early, driving correct structural learning.
\end{itemize}

\vspace{-0.35em}
The \textbf{Sub-Agent} receives a composite reward that balances local execution quality and global outcome alignment:
\begin{equation}
\label{eq:Rsi}
\mathcal{R}_{\mathcal{S}}(o_{\mathcal{S}_i}) = 
\begin{cases}
\beta_1 \cdot r_{\text{format}} + \beta_2 \cdot r_{\text{correct}}^{\text{main}} + \beta_3 \cdot r_{\text{expert}} & \text{valid format} \\
0 & \text{invalid format}
\end{cases}
\end{equation}
where:
\begin{itemize}[leftmargin=25pt]
\vspace{-0.75em}
\setlength\itemsep{-0.25em}
\item $r_{\text{format}}$: Format validation (same zero-reward policy as Main Agent).
\item $r_{\text{correct}}^{\text{main}}$: Main Agent's correctness reward, replicated to Sub-Agent.
\item $r_{\text{expert}}$: LLM-based evaluation of subtask execution quality.
\item $\beta_1$, $\beta_2$, $\beta_3$ are hyperparameters that balance the three reward terms.

\end{itemize}
\vspace{-0.75em}
The coefficients for each component should be settled considering following aspects.
\begin{itemize}[leftmargin=25pt]
\vspace{-0.75em}
\setlength\itemsep{-0.25em}
\item Format weight: Ensures the Sub-Agent outputs valid structures that the Main Agent can parse.
\item Correctness weight: Ensures the Sub-Agent’s output contributes directly to the final result.
\item Expert weight: Assesses the quality of subtask execution, guide the Sub-Agent in tool usage and response.
\end{itemize}
\vspace{-0.75em}

\vspace{-0.25em}
\subsection{Targeted problem}
\vspace{-0.25em}
As shown in Figure \ref{fig:workflow}, multiple sub-agent invocations may occur during one single answering rollout. Therefore, in a batch with multiple training samples, the ratio between the number of main agent and sub-agent trajectories is not fixed and can vary across batches and training steps. To address this issue, we propose M-GRPO, which is detailed with implementation in the following section.

\vspace{-0.65em}
\section{Methodology}
\vspace{-0.30em}
\subsection{M-GRPO}

\vspace{-0.30em}
\textbf{Training objective.} For training the multi-agent system, we adopt Group Relative Policy Optimization (GRPO)~\citep{shao2024deepseekmathpushinglimitsmathematical} and extend it to the multi-agent setting. We refer to this extension as \emph{\textbf{M}ulti-Agent \textbf{GRPO} (M-GRPO)}. The method preserves the group-relative baseline while respecting the hierarchy between $\mathcal{M}$ and $\mathcal{S}$.

\textbf{Group-relative advantages}. For a query $q$, we collect $K$ rollouts, the trajectories and outputs can be denoted: 
\begin{align}
&\{\tau_k=\{\tau_{\mathcal{M}}^{(k)}, \{\tau^{(k)}_{\mathcal{S}_i}\}_{i=1}^{d_k}\}\}_{k=1}^K\\
&\tau^{(k)}_\mathcal{M}=\big\{a^{(k)}_{\mathcal{M},t}, \;s^{(k)}_{\mathcal{M},t}, \;r^{(k)}_{\mathcal{M},t}\big\}_{t=1}^{T_{\mathcal{M}}^{(k)}},\hspace{1em} \tau^{(k)}_{\mathcal{S}_i}=\big\{a^{(k)}_{\mathcal{S}_i,t},\; s^{(k)}_{\mathcal{S}_i,t}, \;r^{(k)}_{\mathcal{S}_i,t}\big\}_{t=1}^{T^{(k)}_{\mathcal{S}_i}}\\[0.3em]
&\{o_k=\{o_{\mathcal{M}}^{(k)}, \{o^{(k)}_{\mathcal{S}_i}\}_{i=1}^{d_k}\}\}_{k=1}^K
\end{align}

For these rollouts, the numbers of main agent $\mathcal{M}$ invocating sub agent $\mathcal{S}$ are $d_1$, $d_2$, ..., $d_K$ respectively, i.e. $i\in\{d_1, d_2$, ..., $d_K\}$. \label{different_number_of_subagent_trajectory} Now we compute centered, normalized returns and advantages. 

For \textbf{main agent} $\mathcal{M}$:
\begin{align}
    &\mu_{q,\mathcal{M}} = \frac{1}{K}\sum_{k=1}^K\mathcal{R}_{\mathcal{M}}(o^{(k)}_{\mathcal{M}}),\hspace{1em}
    \sigma_{q,\mathcal{M}} = \sqrt{\frac{1}{K}\sum_{k=1}^K\left(\mathcal{R}_{\mathcal{M}}(o^{(k)}_{\mathcal{M}})-\mu_{q,\mathcal{M}}\right)^2}\\[0.3em]
    &\hat{A}_{q,\mathcal{M}}^{(k)}=\hat{A}_{q,\mathcal{M},t}^{(k)} = \frac{\mathcal{R}_{\mathcal{M}}(o^{(k)}_{\mathcal{M}})-\mu_{q,\mathcal{M}}}{\sigma_{q,\mathcal{M}}}\hspace{0.5em} \forall\; t\in\{1,2,\cdots,T^{(k)}_\mathcal{M}\}
\end{align}

For \textbf{sub-agent} $\mathcal{S}$, to align the varying invocation counts $d_k$ across $K$ rollouts and enable efficient batch RL training, we propose a batching scheme. Let $d$ denote an approximate upper bound on the number of sub-agent invocations in one rollout, such that
\begin{equation}
    P(d_k\leq d)\approx0.99\hspace{0.75em} \text{$d$ can be selected by a preliminary sanity check}
\end{equation}

For each invocation in one rollout, align the number of subagent trajectories according to the rules:
\begin{itemize}[leftmargin=25pt]
\vspace{-0.75em}
\setlength\itemsep{-0.25em}
    \item If $d_k < d$, randomly duplicate $d-d_k$ trajectories from $\{\tau^{(k)}_{\mathcal{S}_i}\}_{i=1}^{d_k}$ and fill them as a whole batch. For example,  $\tau^{(k)}_{\mathcal{S}_{d_k+1}}=\tau^{(k)}_{\mathcal{S}_j}$ where $j$ is randomly chosen from $\{1,2,...,d_k\}$. 
    \item If $d_k > d$, randomly drop $d_k-d$ trajectories from $\{\tau^{(k)}_{\mathcal{S}_i}\}_{i=1}^{d_k}$.
\end{itemize}
After this adjustment, for $K$ rollouts corresponding to any query $q$, the number of sub-agent invocations will be $d\times K$, and the sub-agent trajectories will become $\{\tau^{(k)}_{\mathcal{S}_i}\}_{i=1}^{d}$. We then follow the analogous computation logic for the main agent $\mathcal{M}$.
\begin{align}
    &\mu_{q,\mathcal{S}}=\frac{1}{d\times K}\sum_{k=1}^K\sum_{i=1}^{d}\mathcal{R}_{\mathcal{S}}(o^{(k)}_{\mathcal{S}_i}),\hspace{1em}
    \sigma_{q,\mathcal{S}}=\sqrt{\frac{1}{d\times K}\sum_{k=1}^K\sum_{i=1}^{d}\left(\mathcal{R}_{\mathcal{S}}(o^{(k)}_{\mathcal{S}_i})-\mu_{q,\mathcal{S}}\right)^2}\\[0.3em]
&\hat{A}^{(k),i}_{q,\mathcal{S}}=\hat{A}^{(k),i}_{q,\mathcal{S},t}=\frac{\mathcal{R}_{\mathcal{S}}(o^{(k)}_{\mathcal{S}_i})-\mu_{q,\mathcal{S}}}{\sigma_{q,\mathcal{S}}}\hspace{0.5em} \forall\; t\in\{1,2,\cdots,T^{(k)}_{\mathcal{S}_i}\}
\end{align}

\textbf{M-GRPO objective}. With defined advantages, the objective functions for main and sub agents are defined as follows.
\begin{align}
    \mathcal{J}_{\text{M-GRPO}}(\theta_{\mathcal{M}}) = &\mathbb{E}_{\{q\sim P(Q),\;\{o^{(k)}_{\mathcal{M}}\}_{k=1}^K\sim\pi_{\theta_\mathcal{M}}(o|q)\}}\notag\\
    &\left[\frac{1}{K}\sum_{k=1}^{K}\min\left(\frac{\pi_{\theta_\mathcal{M}}(o^{(k)}_\mathcal{M}|q)}{\pi_{\theta_{\mathcal{M}_{old}}}(o^{(k)}_\mathcal{M}|q)}\hat{A}_{q,\mathcal{M}}^{(k)},\; \text{clip}\left(\frac{\pi_{\theta_\mathcal{M}}(o^{(k)}_\mathcal{M}|q)}{\pi_{\theta_{\mathcal{M}_{old}}}(o^{(k)}_\mathcal{M}|q)},\; 1-\varepsilon, \; 1+\varepsilon\right)\hat{A}_{q,\mathcal{M}}^{(k)}\right)\right]
\end{align}
\vspace{-0.25em}
where $p(Q)$ is the distribution of the query and $\pi_{\theta_\mathcal{M}}(o^{(k)}_\mathcal{M}|q)$ is the likelihood of the final answer given the query $q$.
\vspace{-0.25em}
\begin{align}
    \pi_{\theta_\mathcal{M}}(o^{(k)}_\mathcal{M}|q)=\prod_{t=1}^{T^{(k)}_\mathcal{M}}\pi_{\theta_\mathcal{M}}(a^{(k)}_{\mathcal{M},t}|s^{(k)}_{\mathcal{M},t})
\end{align}

The sub-agent objective sums over all its invocations:
\begin{align}
    \mathcal{J}_{\text{M-GRPO}}(\theta_{\mathcal{S}}) = &\mathbb{E}_{\{q\sim P(Q),\;\{o^{(k)}_{\mathcal{S}_1},...,o^{(k)}_{\mathcal{S}_{d}}\}_{k=1}^K\sim\pi_{\theta_\mathcal{M}}(o|q)\}}\notag\\
    &\left[\frac{1}{dK}\sum_{k=1}^{K}\sum_{i=1}^{d}\min\left(\frac{\pi_{\theta_\mathcal{S}}(o^{(k)}_{\mathcal{S}_i}|q)}{\pi_{\theta_{\mathcal{S}_{old}}}(o^{(k)}_{\mathcal{S}_i}|q)}\hat{A}_{q,\mathcal{S}}^{(k),i},\; \text{clip}\left(\frac{\pi_{\theta_\mathcal{S}}(o^{(k)}_{\mathcal{S}_i}|q)}{\pi_{\theta_{\mathcal{S}_{old}}}(o^{(k)}_{\mathcal{S}_i}|q)},\; 1-\varepsilon, \; 1+\varepsilon\right)\hat{A}_{q,\mathcal{S}}^{(k),i}\right)\right]
\end{align}
Similarly, $\pi_{\theta_\mathcal{S}}(o^{(k)}_{\mathcal{S}_i}|q)$ is the likelihood of the answer to the i-th sub-agent invocation given the query $q$.
\begin{align}
    \pi_{\theta_\mathcal{S}}(o^{(k)}_{\mathcal{S}_i}|q)=\prod_{t=1}^{T^{(k)}_{\mathcal{S}_i}}\pi_{\theta_{\mathcal{S}_i}}(a^{(k)}_{\mathcal{S}_i,t}|s^{(k)}_{\mathcal{S}_i,t})
\end{align}

We maximize $\mathcal{J}_{\text{M-GRPO}}(\theta_{\mathcal{M}})$ and $\mathcal{J}_{\text{M-GRPO}}(\theta_{\mathcal{S}})$ separately for distinct LLMs integrated in main and sub agents respectively. Specifically:
\begin{align}
    &\hat{\theta}_{\mathcal{M}}=\arg\max_{\theta_{\mathcal{M}}}\mathcal{J}_{\text{M-GRPO}}(\theta_{\mathcal{M}})\\
    &\hat{\theta}_{\mathcal{S}}=\arg\max_{\theta_{\mathcal{S}}}\mathcal{J}_{\text{M-GRPO}}(\theta_{\mathcal{S}})
\end{align}

\subsection{Implementation}
\begin{figure}[ht]
    \centering
    \includegraphics[width=0.95\linewidth]{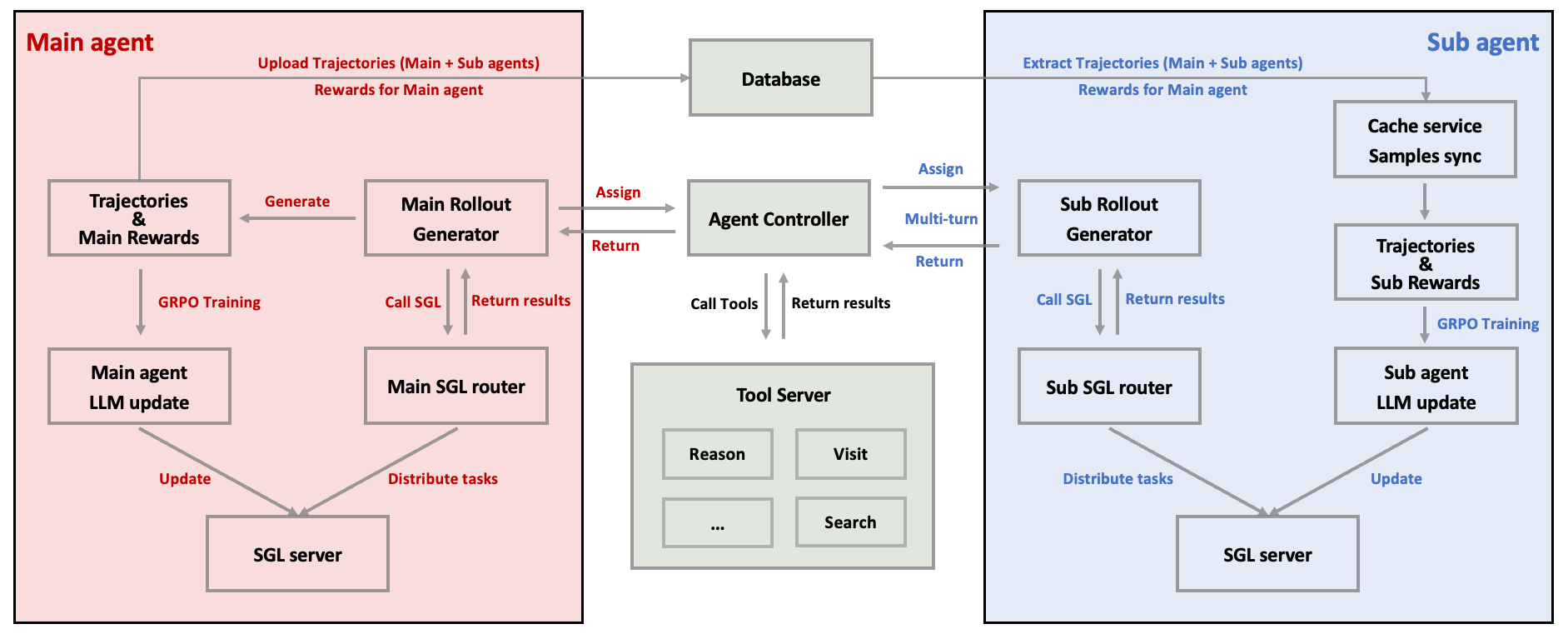}
    \caption{\textbf{Workflow of the decoupled two-agent architecture with M-GRPO.} The Main agent (left) and Sub agent (right) each generate rollouts via their SGL router/server. Main agent logs trajectories and rewards to a shared Database. Sub agent extracts the required rewards from the database and calculates its own rewards for training. A central Agent Controller (middle) coordinates multi-turn interactions, assigns subtasks to the sub agent, and aggregates returned results. Tool calls (reason/search/visit) are executed through a Tool Server. The sub-agent side maintains a cache for sample synchronization. Arrows indicate data and control flow.}
    \label{fig:workflow_detail}
\end{figure}

\begin{figure}[ht]
    \centering
    \includegraphics[width=0.95\linewidth]{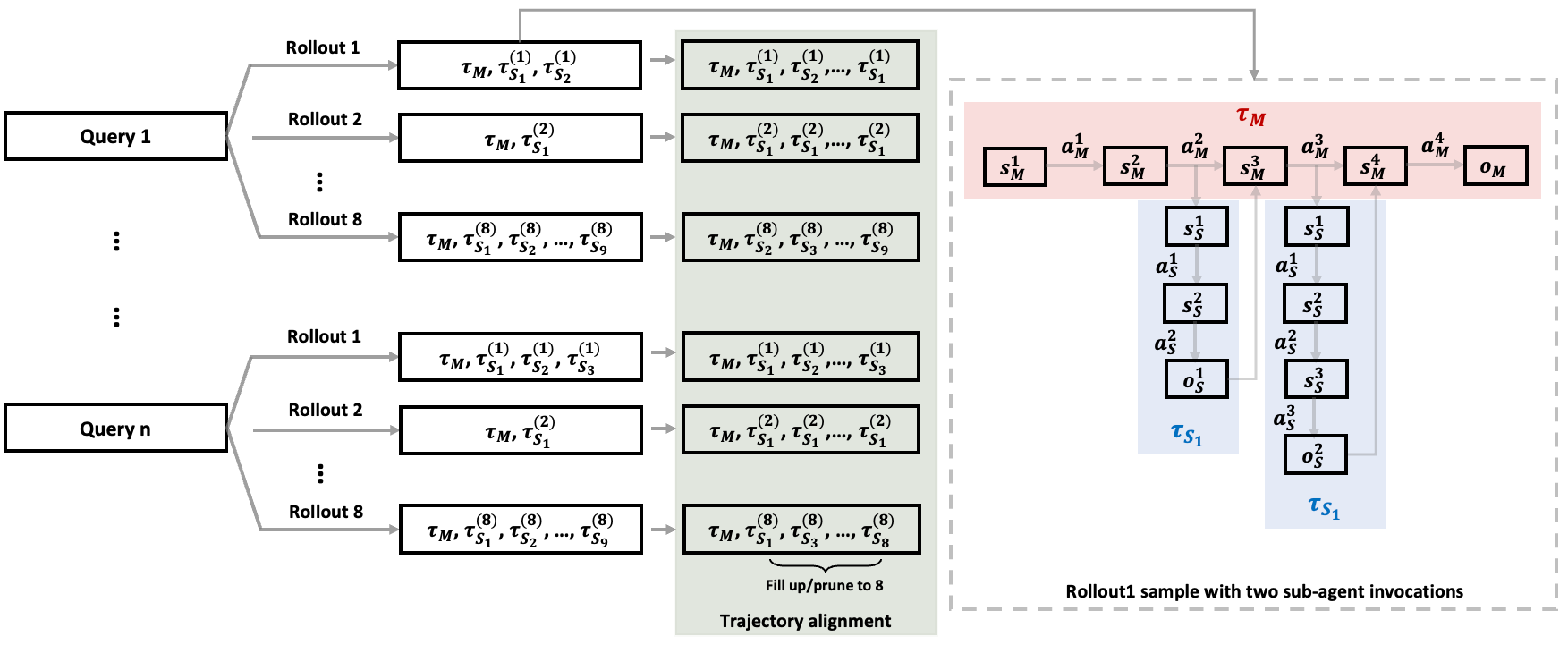}
    \caption{\textbf{Trajectory alignment for batch training with variable sub-agent invocations.}
  For each query, we sample $K$ rollouts. Every rollout yields one main-trajectory \(\tau_{\mathcal{M}}\) and a variable number of sub-agent trajectories \(\{\tau_{\mathcal{S}_i}\}\).
  Because the number of sub-invocations \(d_k\) differs across rollouts, we fix a target $d$ (e.g., \(8\)) and randomly duplicate or drop \(\tau_{\mathcal{S}}\) samples so that each batch contains a consistent count of \(\tau_{\mathcal{S}}\) (while keeping a fixed number of \(\tau_{\mathcal{M}}\), e.g., $8$).
  This alignment produces uniform tensor shapes for policy-gradient updates.}
    \label{fig:trajectory_alignment}
\end{figure}

We deploy the main agent and the sub-agent on separate rollout servers, server $M$ and server $S$. For each query in a batch, we run $8$ rollouts and select $d$ to be $8$.
Following prior work \citep{mrlx2025}, we update the main agent on server $M$ with its rewards and write its rewards to a shared database. We compute the log likelihood on batches containing $\tau_\mathcal{M}$ and $\tau_\mathcal{S}$ separately using $\pi_{\theta_\mathcal{M}}$, $\pi_{\theta_{\mathcal{M}_{old}}}$ and $\pi_{\theta_\mathcal{M}}$, $\pi_{\theta_{\mathcal{S}_{old}}}$. Server $S$ then reads the required rewards from the database, computes the rewards for sub-agents and updates the sub-agent accordingly. The illustration of the subagent trajectory alignment process and training on different servers is shown in Figure \ref{fig:workflow_detail}, \ref{fig:trajectory_alignment}. As for tools, main agent has the access to a reasoning tool which is able to solve logical math problems. Sub agents can invoke visit and search tools for desired knowledge.

\vspace{-0.5em}
\section{Experiments}
\vspace{-0.30em}

We evaluate our multi-agent training architecture on domain-specific and general-purpose benchmarks. In this paper, we focus on the scenario that the multi-agent system tries to find the answer for a given query by a multi-turn deep search loop. Performance is measured by (i) answer correctness and (ii) adherence to the required output format.

\vspace{-0.30em}
\subsection{Experimental Setup}

\textbf{Curriculum learning.} We adopt a two-stage reinforcement learning curriculum ~\citep{narvekar2020curriculumlearningreinforcementlearning}. In stage 1, we train the model on a simple dataset to enable rapid learning of basic answering formats. In stage 2, we introduce more complex problems that encourage collaborative problem-solving within the multi-agent framework.

\vspace{-0.15em}
\textbf{Training data.} The training datasets for both stages are constructed separately. For stage 1, we use the data synthesis approach from \citet{yu2025medresearcherr1expertlevelmedicaldeep}, but with simpler graphs containing fewer nodes and edges to facilitate quick format learning. For stage 2, we adopt the same method but by building graphs with more nodes and edges. This helps to create more challenging datasets. We also select queries that achieve low success rates in the same multi-agent system with untrained LLMs, thereby focusing on difficult collaborative problem-solving scenarios.

\vspace{-0.15em}
\textbf{Benchmarks.} We evaluate our trained MAS on three benchmarks.

\begin{itemize}[leftmargin=20pt, itemsep=-0.25em]
\vspace{-0.5em}
\item \textbf{GAIA} \citep{mialon2023gaiabenchmarkgeneralai}. Tests real-world assistant capabilities on multi-modal tasks involving tools, web search, and multi-step reasoning. It emphasizes human-simple yet AI-hard skills, such as reading comprehension, logical reasoning, and effective tool use in realistic settings.
\item \textbf{XBench-DeepSearch} \citep{chen2025xbenchtrackingagentsproductivity}. Assesses tool use across diverse open-domain scenarios, including fact-checking, comparative analysis, browsing-based reasoning, and complex information synthesis. It provides a broad measure of real-world problem solving.
\item \textbf{WebWalkerQA} \citep{wu2025webwalkerbenchmarkingllmsweb}. Evaluates web navigation and information extraction capabilities through multi-step browsing tasks.
\end{itemize}

\vspace{-0.75em}
\textbf{Baselines}. To validate the performance improvement of updating both main and sub agents together with M-GRPO, we compare our methods with two baselines. To better illustrate the role of M-GRPO in training, we initialize all baselines from the same stage 1 training checkpoint.

\begin{itemize}[leftmargin=20pt, itemsep=-0.25em]
\vspace{-0.75em}
\item \textbf{Multi-agent system with fixed sub-agent}. A multi-agent system with the same architecture, where the main agent is updated using common GRPO, while the parameters of the sub-agent are frozen. To compare performance on answering rather than formatting, we freeze the sub-agent parameters exclusively in Stage 2.
\end{itemize}

\vspace{-0.75em}
\textbf{Model and infrastructure.} We initialize both single-agent and multi-agent systems with Qwen3-30B-A3B \citep{qwen3technicalreport}. All training is conducted using the SLIME framework \citep{slime_github}. The single-agent baseline is trained on 32 H800 GPUs, while the multi-agent system uses $2\times32$ H800 GPUs, with one cluster for the main agent ($\mathcal{M}$) and one for sub-agent ($\mathcal{S}$).

\textbf{Training algorithms.} For the single-agent baseline, we provide access to all tools available to the sub-agent $\mathcal{S}$ in the multi-agent system, while keeping other configurations (model architecture, hyperparameters, etc.) identical to the main agent $\mathcal{M}$. We apply GRPO with the format reward weight $\alpha_1=0.1$ and $\alpha_2=0.9$ in Eq.~\eqref{eq:Rm}. In the multi-agent system, the main agent $\mathcal{M}$ uses M-GRPO with the reward structure specified in Eq.~\eqref{eq:Rm} and the same $\alpha$ values. The sub-agent $\mathcal{S}$ receives a composite reward as defined in Eq.~\eqref{eq:Rsi} with $\beta_1=0.1$, $\beta_2=0.4$ and $\beta_3=0.5$. The 0.4/0.5 balance is crucial. Too much weight on global outcome creates noisy gradients (Sub-Agent performed well but Main Agent failed elsewhere), while too much weight on local quality allows Sub-Agents to optimize for metrics that do not contribute to final success.

\subsection{Stage 1: Format learning with simple data}
\begin{figure}[ht]
    \centering
    \includegraphics[width=0.65\linewidth]{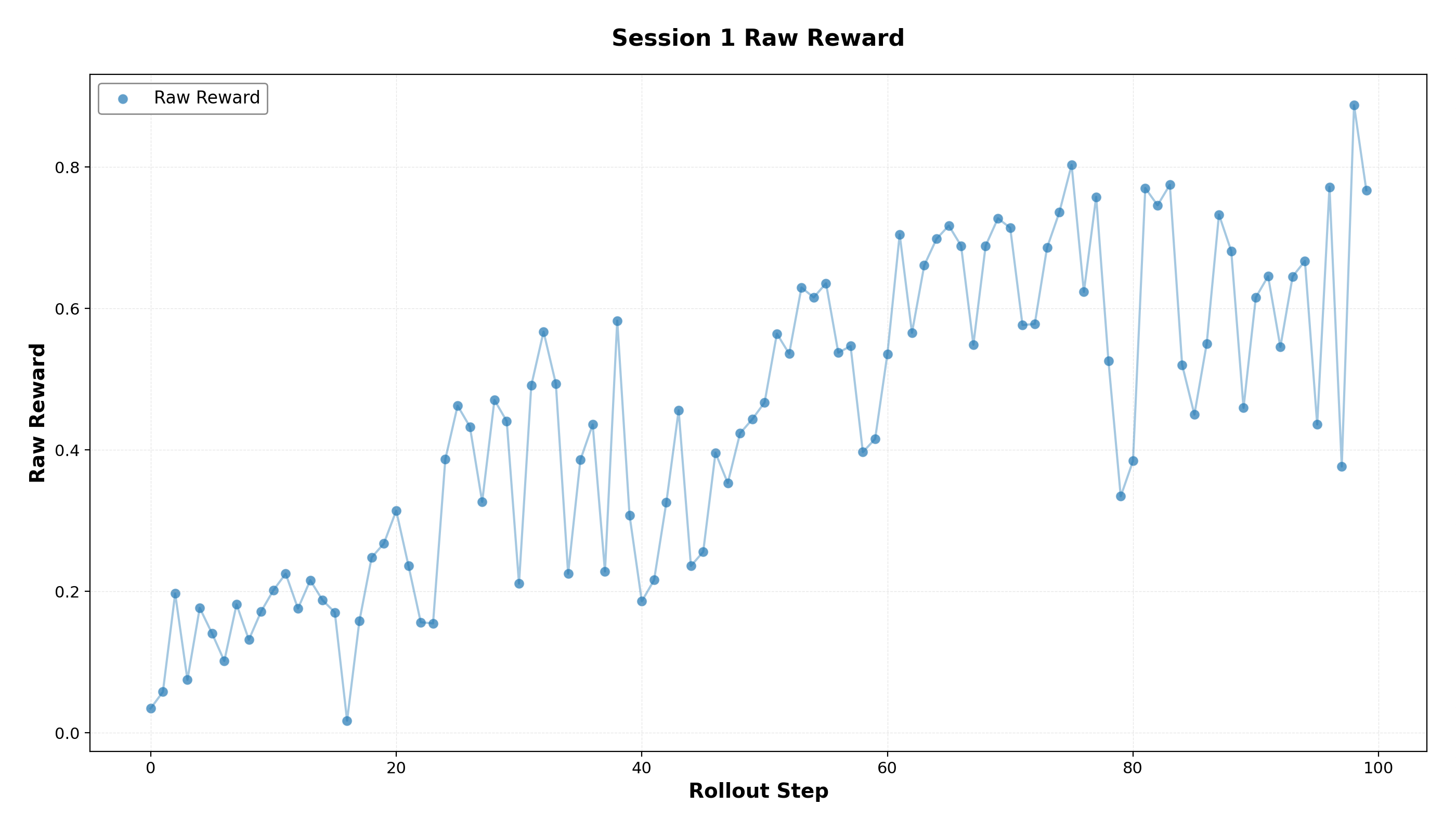}
    \vspace{0.25em}
    \caption{\textbf{Reward curve during Stage 1 RL training on simple data}. The system shows stable improvement from zero to high rewards, demonstrating effective format acquisition.}
    \label{fig:reward_stage1}
\end{figure}

\vspace{-0.35em}
Figure~\ref{fig:reward_stage1} shows the reward curve during stage 1 RL training. The reward trajectory during stage 1 demonstrates a stable learning with rewards rising steadily from zero to a high plateau. This validates that our curriculum design enables rapid acquisition of proper output formatting with a simple QA dataset in the first stage. It also establishes a foundation for subsequent complex collaborative learning.

\vspace{-0.35em}

\subsection{Stage 2: Learning multi-agent collaboration on complex tasks}
\vspace{-0.35em}
After establishing format competence in stage 1, we proceed to stage 2 RL training using challenging problems and designed M-GRPO. The goal is to elicit and strengthen collaborative problem-solving behaviors between $\mathcal{M}$ and $\mathcal{S}$ under complex task demands.

\subsubsection{Main results: benchmark performance}

\begin{figure}[htbp]
    \centering
    \includegraphics[width=0.95\linewidth]{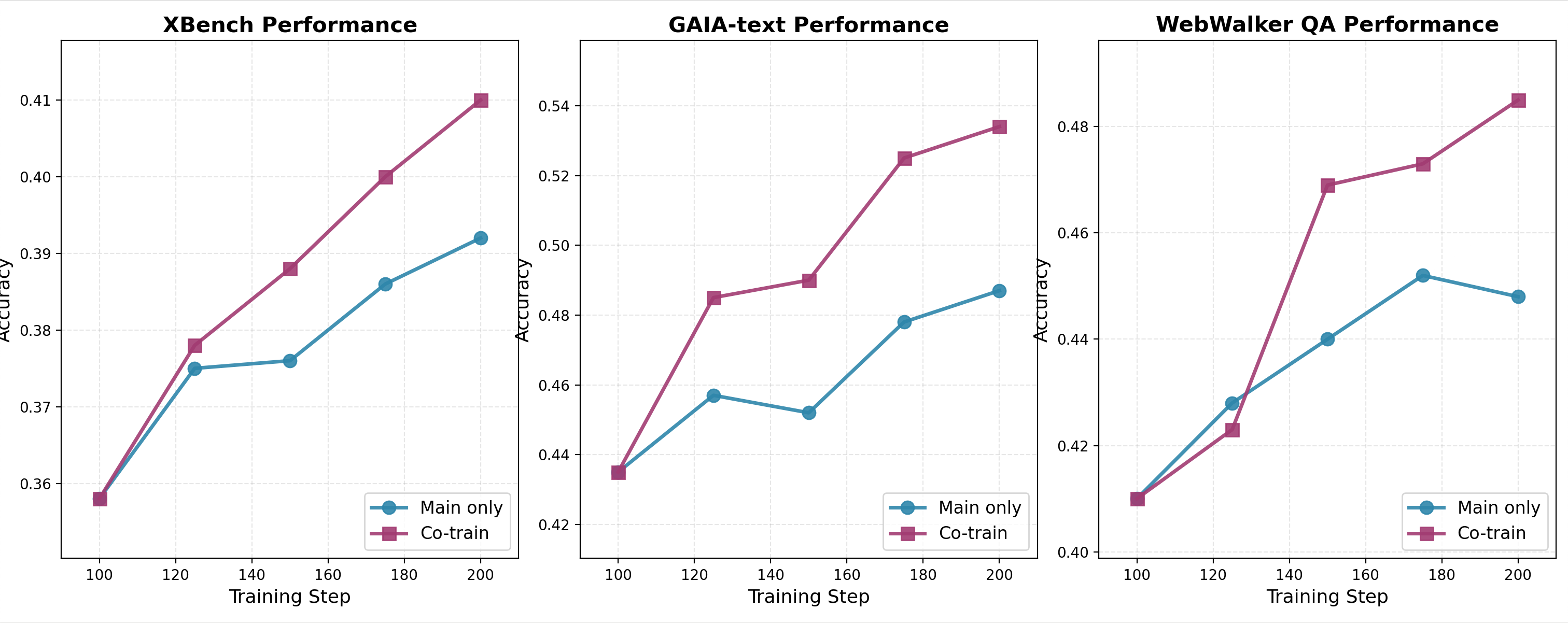}
    \vspace{0.25em}
    \caption{\textbf{Benchmark performance during Stage 2 training (averaged over three runs)}. Co-training both agents (blue) consistently outperforms main-only training (purple) across GAIA, XBench-DeepSearch, and WebWalkerQA benchmarks, demonstrating that learned collaborative behaviors transfer to diverse real-world tasks.}
    \label{fig:inference_results}
\end{figure}

To assess whether the collaborative capabilities learned during stage 2 RL training generalize to real-world tasks, we evaluate checkpoints on three benchmarks throughout the stage 2 training process. Figure~\ref{fig:inference_results} presents performance trajectories on GAIA, XBench-DeepSearch, and WebWalkerQA, evaluated every 25 training steps. We compare co-training (both agents updated) versus main-only training (sub-agent fixed) and single-agent training. All reported results are averaged over three independent inference runs to ensure reliability.

The results show that co-training yields consistent improvements over main-only training across all three benchmarks throughout stage 2. This demonstrates that the collaborative problem-solving capabilities elicited by M-GRPO on challenging training data effectively transfer to diverse real-world evaluation scenarios.

\subsubsection{Ablation study: training configurations}

To understand the source of the performance gains observed in Figure~\ref{fig:inference_results} and validate the contribution of different components in our framework, we conduct ablation studies comparing three training configurations during stage 2:

\begin{enumerate}[leftmargin=20pt]
\vspace{-0.5em}
\setlength\itemsep{-0.1em}
\item \textbf{Multi-agent co-training}: Both $\mathcal{M}$ and $\mathcal{S}$ are trained jointly using M-GRPO (our full method).
\item \textbf{Multi-agent main-only}: Only $\mathcal{M}$ is trained while $\mathcal{S}$ remains fixed (the orange curves in Figure~\ref{fig:inference_results}).
\item \textbf{Single-agent baseline}: A single agent handles both planning and tool execution with access to all tools available to $\mathcal{S}$, maintaining identical model architecture and hyperparameters as $\mathcal{M}$. This baseline is trained with a separate configuration including its own format warm-up phase.
\end{enumerate}

\begin{figure}[htbp]
    \centering
    \includegraphics[width=0.65\linewidth]{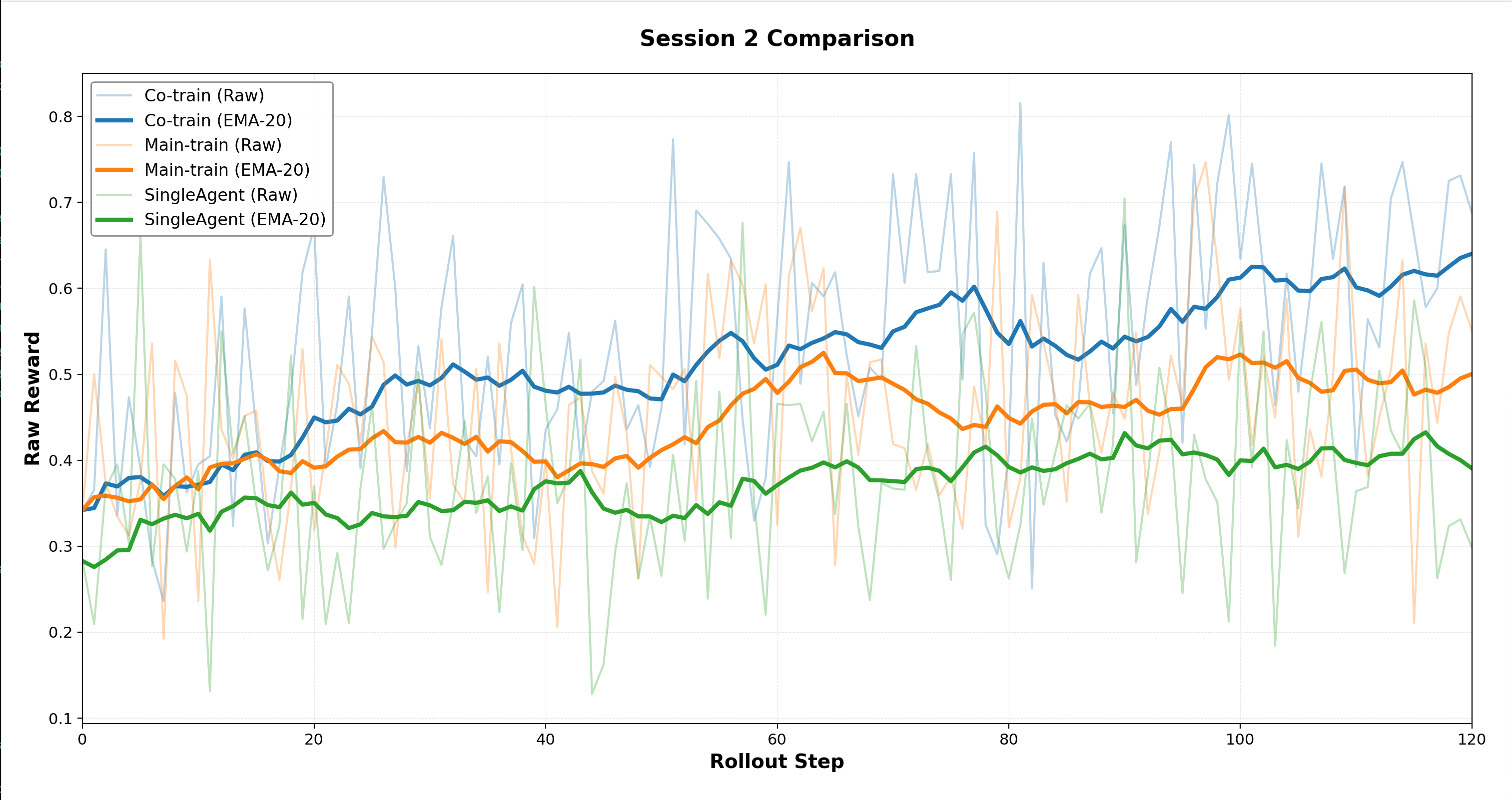}
    \vspace{0.25em}
    \caption{\textbf{Stage 2 RL learning curves on challenging data comparing three training configurations}. Raw rewards and EMA-smoothed trends are shown. Co-training both agents achieves the highest rewards, followed by main-only training, with single-agent baseline showing the lowest performance.}
    \label{fig:stage2_compare}
\end{figure}

For fair comparison, all multi-agent configurations initialize from the stage 1 checkpoint. Figure~\ref{fig:stage2_compare} shows the stage 2 learning curves. Since the raw reward curves exhibit high variance, we overlay exponential moving average (EMA) smoothed curves to better visualize the training trends.

The results demonstrate a clear performance hierarchy: \textbf{co-training $>$ main-only $>$ single-agent}. This validates that (1) the multi-agent architecture itself provides benefits over single-agent systems when facing complex tasks, and (2) joint optimization of both agents through M-GRPO further improves collaborative problem-solving compared to training only the main agent.

\subsubsection{Ablation study: trajectory synchronization}

To maintain more on-policy training dynamics, we introduce a sample-level synchronization strategy as described in Section~\ref{different_number_of_subagent_trajectory}. This mechanism aligns trajectories across agents by fixing the number of sub-agent invocations to a target value $d$ through duplication or dropping of sub-trajectories. To validate the effectiveness of this design choice, we compare two implementations during stage 2:

\begin{enumerate}[leftmargin=20pt]
\vspace{-0.5em}
\setlength\itemsep{-0.1em}
\item \textbf{Without synchronization}: An earlier implementation that uses off-policy data without considering sample-level alignment across agents.
\item \textbf{With synchronization}: Our current implementation that enforces trajectory alignment to maintain more consistent policy-data correspondence.
\end{enumerate}

\begin{figure}[htbp]
    \centering
    \includegraphics[width=0.65\linewidth]{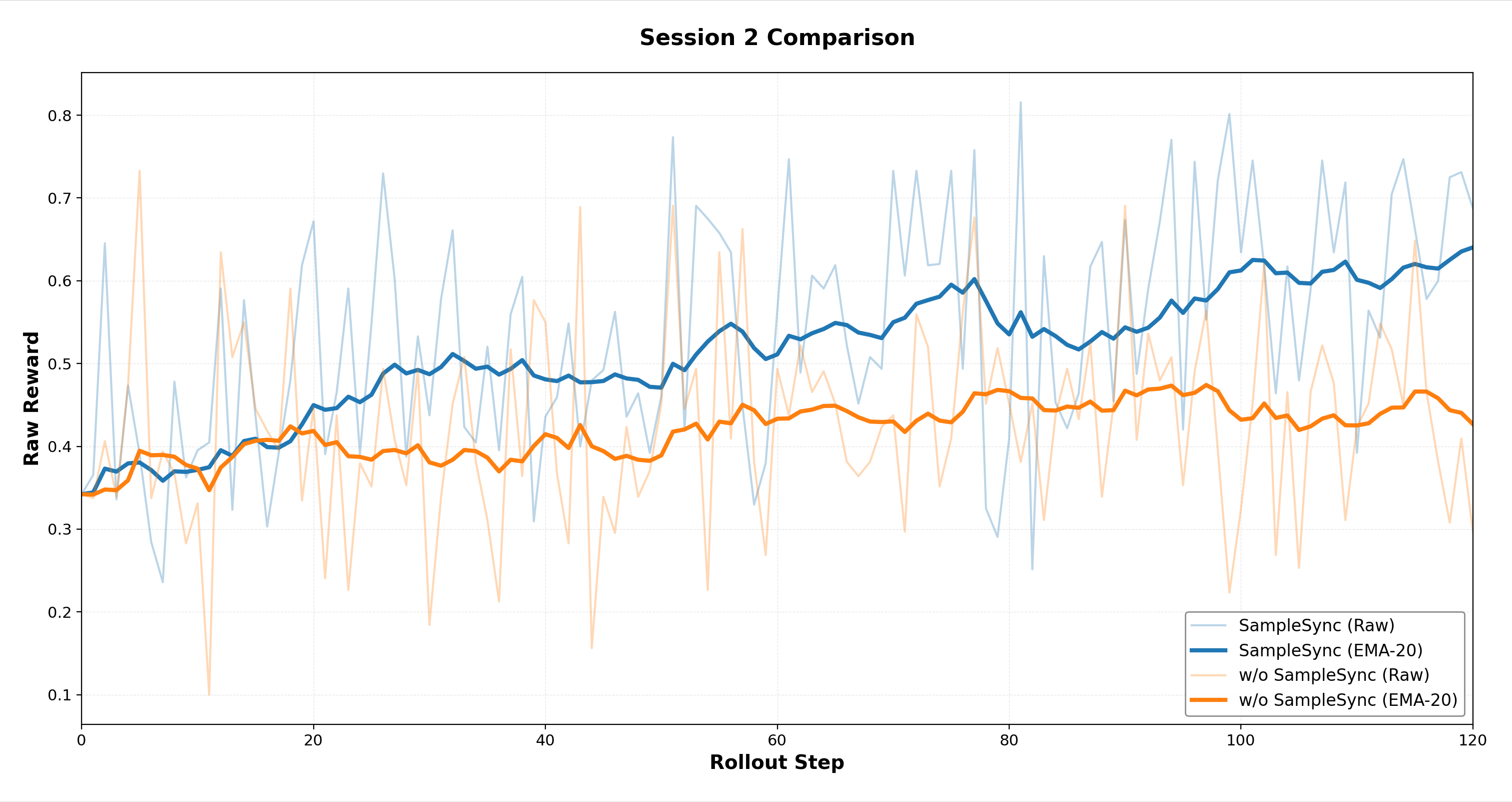}
    \vspace{0.25em}
    \caption{\textbf{Stage 2 RL learning curves comparing implementations with and without trajectory synchronization}. Raw rewards (lighter lines) and EMA-smoothed trends (darker lines) are shown. The synchronized version (blue) outperforms the unsynchronized version (orange), validating the benefit of maintaining more on-policy training dynamics.}
    \label{fig:stage2_sync}
\end{figure}

Figure~\ref{fig:stage2_sync} shows the stage 2 learning curves for both implementations, with EMA smoothing applied to highlight trends. The synchronized version achieves better performance, demonstrating that sample-level synchronization between agents improves training stability and collaborative learning effectiveness by keeping the training process closer to on-policy.

\subsection{Summary}

Our experiments demonstrate several key findings:

\begin{itemize}[leftmargin=20pt]
\vspace{-0.5em}
\setlength\itemsep{-0.1em}
\item The two-stage RL curriculum (stage 1 on simple data for format learning, stage 2 on challenging data for collaborative capability development) enables stable training progression (Figure~\ref{fig:reward_stage1}).
\item Co-training both agents consistently outperforms main-only training across multiple real-world benchmarks throughout stage 2, demonstrating effective transfer of learned collaborative behaviors (Figure~\ref{fig:inference_results}).
\item During stage 2 RL training, the performance hierarchy co-training $>$ main-only $>$ single-agent validates both the multi-agent architecture and the M-GRPO joint optimization approach (Figure~\ref{fig:stage2_compare}).
\item Trajectory synchronization improves training stability and collaborative learning effectiveness during stage 2 by maintaining more on-policy training dynamics (Figure~\ref{fig:stage2_sync}).
\end{itemize}

These results validate M-GRPO as an effective framework for training collaborative LLM agents on complex, tool-augmented tasks.

\section{Conclusion}

In this paper, we introduce M-GRPO, an effective reinforcement learning framework designed for training multi-agent systems that utilize LLMs. Our approach leverages hierarchical credit assignment and trajectory alignment to address key challenges in vertical multi-agent architectures, including imbalanced rollout numbers and complex cross-server gradient flow. Through empirical evaluation, we demonstrate that M-GRPO outperforms traditional single-agent systems and fixed sub-agent configurations on real-world benchmarks, such as GAIA, XBench and WebWalker QA.

The findings highlight the importance of multi-agent collaboration in complex tasks, where M-GRPO effectively fosters task-specific expertise among sub-agents while maintaining global alignment with the main agent’s goals. Our experimental results validate that joint optimization of both main and sub-agents leads to superior performance in long-horizon planning and tool-augmented reasoning.

\bibliographystyle{plainnat}
\bibliography{ref}

\newpage
\appendix

\section{Appendix}
In appendix, we present several case studies to show that the multi-agent system can learn targeted functions.
\subsection{Tool call}
After training with M-GRPO, muti-agent system (MAS) learns how to select the optimal tools for question solution.
\begin{tcolorbox}[colback=gray!10!white, colframe=gray!80!black, title=\textbf{Question}, left=0.5em]
A standard Rubik’s cube has been broken into cubes making up its sides. The cubes are jumbled, and one is removed. There are 6 cubes with one colored face, 12 edge cubes with two colored faces, and 8 corner cubes with three colored faces. All blue cubes have been found. All cubes directly left, right, above, and below the orange center cube have been found, along with the center cube. The green corners have all been found, along with all green that borders yellow. For all orange cubes found, the opposite face’s cubes have been found. The removed cube has two colors on its faces. What are they? Answer using a comma separated list, with the colors ordered alphabetically.
\end{tcolorbox}

The correct answer should be "green, white." Before training, MAS with the same vertical architecture produces the following outputs. The answer given before training is incorrect, whereas after training, the MAS can correctly answer the question. We explore the reasons for this discrepancy in the MAS's thinking process.

\begin{minipage}[t]{0.49\textwidth}
    \begin{tcolorbox}[colback=red!5!white, colframe=red!80!black, title=\textbf{Answer before training}, left=0.5em]
    "prediction": "Green,Orange"
\end{tcolorbox}
\end{minipage}\hfill
\begin{minipage}[t]{0.49\textwidth}
    \begin{tcolorbox}[colback=green!5!white, colframe=green!80!black, title=\textbf{Answer after training}, left=0.5em]
    "prediction": "green, white"
\end{tcolorbox}
\end{minipage}

The thinking processes differ before and after training in a MAS, as the system learns how to select the appropriate tools for a given question. The following outlines the initial responses upon receiving a question. Before training, the MAS invokes the "browser" tool, but after training, it uses the "reasoner" tool. In this particular case, where logical derivation is required, the "reasoner" tool is the more appropriate choice. 

For the whole answering process, before training, the main agent in the MAS does not invoke the "reasoner" tool at all, instead calling the sub-agent twice to handle searching and browsing tasks. However, after training, the MAS uses the "reasoner" tool twice and calls the sub-agent only once, primarily to verify basic information about the Rubik's cube. This example demonstrates how the training process helps the system choose the most suitable tool for a given task.

\begin{paracol}{2}
\begin{tcolorbox}[colback=red!5!white, colframe=red!80!black, title=\textbf{Thinking before training}, left=0.25em, breakable, enhanced]
role: assistant
    
content: <thinking>
To solve this puzzle, I need to determine the two colors on the removed cube. Let's break down the information:

\vspace{-0.35em}
\begin{enumerate}[leftmargin=1.5em]
\setlength\itemsep{-0.25em}
    \item A standard Rubik’s cube has 27 smaller cubes: 6 center cubes (1 color), 12 edge cubes (2 colors), 8 corner cubes (3 colors), and 1 central cube (no color).
    \item  The removed cube has 2 colors (it's an edge cube).
    \item All blue cubes have been found.
    \item All cubes adjacent to the orange center cube (left, right, above, below) have been found, along with the orange center cube.
    \item All green corner cubes have been found, and all green pieces that border yellow have been found.
\end{enumerate}
Given these constraints, I need to:
\vspace{-0.35em}
\begin{itemize}[leftmargin=1.5em]
\setlength\itemsep{-0.25em}
    \item[-] Identify which edge cubes could be missing.
    \item[-] Use the constraints to eliminate possibilities.
    \item[-] Deduce the two colors of the removed cube.
\end{itemize}
\vspace{-0.5em}
I'll start by searching for the standard color configuration of a Rubik’s cube and the position of the center cubes to understand the relationships between colors.
</thinking>

<tool\_call>\{"name": \colorbox{orange}{"browser"}, "arguments": \{"task": "standard Rubik's cube color arrangement center faces", "context": ""\}\}</tool\_call>
\end{tcolorbox}

\switchcolumn
\begin{tcolorbox}[colback=green!5!white, colframe=green!80!black, title=\textbf{Thinking process after training}, left=0.25em, breakable, enhanced]
role: assistant
    
content: <thinking>
This is a complex logic puzzle about a Rubik's cube. I need to carefully analyze the information given and use logical reasoning to determine which cube was removed and what colors are on its faces.

Let me break down what I know:
\vspace{-0.35em}
\begin{enumerate}[leftmargin=1.5em]
\setlength\itemsep{-0.25em}
  \item A standard Rubik’s cube has 27 smaller cubes total (3x3x3).
  \item The cube has been broken apart and one cube is missing.
  \item There are 6 cubes with one colored face (center pieces).
  \item 12 edge cubes with two colored faces.
  \item 8 corner cubes with three colored faces.
  \item Total: 6 + 12 + 8 = 26 cubes, so one is missing (which matches the puzzle).
\end{enumerate}
\vspace{-0.35em}
Given information:
\begin{itemize}[leftmargin=1.5em]
\setlength\itemsep{-0.25em}
  \item All blue cubes have been found.
  \item All cubes directly left, right, above, and below the orange center cube have been found....
\end{itemize}
\vspace{-0.35em}
I need to use logical reasoning to determine which cube is missing and what colors it has. Let me start by analyzing the implications of each clue. </thinking>

<tool\_call>\{"name": \colorbox{orange}{"reasoner"}, "arguments": \{"problem": "A standard Rubik's cube has 27 smaller cubes total ... What are the two colors on the missing cube? Answer with colors in alphabetical order separated by commas."\}</tool\_call>

\end{tcolorbox}
\end{paracol}


\subsection{Split and distribute tasks}

An important ability that the main agent learns from training with M-GRPO is to properly distribute tasks for sub agent. This progress greatly improves the performance of the sub agent and the results from tools. An example is shown below.

\begin{tcolorbox}[colback=gray!10!white, colframe=gray!80!black, title=\textbf{Question}, left=0.5em]
I’m researching species that became invasive after people who kept them as pets released them. There’s a certain species of fish that was popularized as a pet by being the main character of the movie Finding Nemo. According to the USGS, where was this fish found as a nonnative species, before the year 2020? I need the answer formatted as the five-digit zip codes of the places the species was found, separated by commas if there is more than one place.
\end{tcolorbox}

The correct answer to the question is "34689". Before training, the MAS produces an incorrect prediction of "33755". After training, however, it returns the correct answer. The comparison of reasoning processes reveals that the incorrect prediction is primarily caused by ambiguous task assignment to the sub-agent.

Before training, the sub-agent's search task fails to include a critical detail from the question: "became invasive after being released by pet owners." Instead, it retrieves general information about the invasive species Ocellaris Clownfish, which is not directly relevant. In contrast, after training, the main agent explicitly instructs the sub-agent to search for species that "became invasive after being released by pet owners" and provides a clear search scope, including references to "USGS data" and "found as nonnative before 2020."

This case demonstrates that the main agent develops stronger reasoning capabilities after training. It can identify missing but essential information in the question and assign more accurate and context-aware tasks to sub-agents.

\begin{paracol}{2}
\begin{tcolorbox}[colback=red!5!white, colframe=red!80!black, title=\textbf{Thinking before training}, left=0.25em, breakable, enhanced]
role: assistant
\vspace{0.25em}

content: <thinking>
I need to identify the species of fish from the movie "Finding Nemo" that became invasive after being released by pet owners. The fish is the Clownfish, specifically the Ocellaris Clownfish (Amphiprion ocellaris). I will search for information from the USGS (United States Geological Survey) on where this species was found as a nonnative species before 2020, specifically focusing on five-digit zip codes.
</thinking>

\vspace{0.25em}
<tool\_call>\{"name": "browser", "arguments": \{"task": \colorbox{orange}{"Find USGS data
on nonnative Ocellaris} \colorbox{orange}{Clownfish (Amphiprion ocellaris) occurrences in the} \colorbox{orange}{USA before 2020, including zip codes.}", "context": "Ocellaris Clownfish, popularized by the movie Finding Nemo, has been reported as an invasive species in the USA."\}\}\}</tool\_call>
\end{tcolorbox}

\switchcolumn
\begin{tcolorbox}[colback=green!5!white, colframe=green!80!black, title=\textbf{Thinking process after training}, left=0.25em, breakable, enhanced]
role: assistant
    
content: <thinking>

I need to research information about the fish species from "Finding Nemo" that became invasive after being released by pet owners. The question specifically asks for:
\begin{enumerate}[leftmargin=1.5em, itemsep=-0.25em]
\vspace{-0.5em}
    \item The fish species that was popularized by "Finding Nemo".
    \item Where this species was found as a nonnative species before 2020 according to USGS.
    \item The answer should be formatted as five-digit zip codes. 
\end{enumerate}

\vspace{-0.5em}

Let me start by researching the fish species from Finding Nemo and its invasive status.
</thinking>

\vspace{0.25em}

<tool\_call>\{"name": "browser", "arguments": \{"task": \colorbox{orange}{"Search for information about the fish} \colorbox{orange}{species from Finding Nemo that became invasive} \colorbox{orange}{after being released by pet owners, specifically look-} \colorbox{orange}{ing for USGS data on where this species was found} \colorbox{orange}{as nonnative before 2020"}\}\}</tool\_call>
\end{tcolorbox}
\end{paracol}

\end{document}